\documentclass[journal]{IEEEtran}
\usepackage{subfigure}
\usepackage{amsmath}
\usepackage{graphicx}
\usepackage{cite}
\usepackage{multirow}
\usepackage{amssymb}

\ifCLASSINFOpdf
\else
\fi
\begin{document}
%
\title{DAPnet: A Double Self-attention Convolutional Network for Point Cloud Semantic Labeling}
%
%
%

\author{Li~Chen,
        Weiye~Chen,
        Zewei~Xu,
        Haozhe~Huang,
        Shaowen~Wang,
        Qing~Zhu,
        Haifeng~Li,~\IEEEmembership{Member,~IEEE}

\thanks{This work was supported by the National Natural Science Foundation of China (grant numbers 41871364, 61773360, 41861048, 41871276 and 41871302). Corresponding author: Haifeng Li Email: \mbox{lihaifeng@csu.edu.cn}.}

    \thanks{Li Chen, Haozhe Huang, and Haifeng Li are with the School of Geosciences and Info-Physics, 
    Central South University, South Lushan Road, Changsha, 410083, China.
    Email: \mbox{\{vchenli,hz\_huang,lihaifeng\}@csu.edu.cn}.}

\thanks{Weiye Chen, Zewei Xu, and Shaowen Wang are with the Department of Geography and Geographic Information Science, University of Illinois at Urbana-Champaign, USA.; CyberGIS Center for Advanced Digital and Spatial Studies, University of Illinois at Urbana-Champaign, USA. Email: \mbox{\{weiyec2,zeweixu2,shaowen\}@illinois.edu}}

\thanks{Qing Zhu is with the Department of Geosciences and Environmental Engineering, the Southwest Jiaotong University, Chengdu, China. Email: \mbox{zhuq66@263.net}}

\thanks{Citation: L. Chen, W. Chen, Z. Xu, H. Huang, S. Wang, Q. Zhu, H. Li. DAPnet: A Double Self-attention Convolutional Network for Point Cloud Semantic Labeling. IEEE Journal of Selected Topics in Applied Earth Observations and Remote Sensing. 2021. DOI: 10.1109/JSTARS.2021.3113047}

}

%
%

\markboth{Accept by IEEE Journal of Selected Topics in Applied Earth Observations and Remote Sensing, 2021}%
{Shell \MakeLowercase{\textit{Li Chen et al.}}: DAPnet: A Double Self-attention Convolutional Network for Point Cloud Semantic Labeling}
%



\maketitle

\begin{abstract}
Airborne Laser Scanning (ALS) point clouds have complex structures, and their 3D semantic labeling has been a challenging task. It has three problems: (1) the difficulty of classifying point clouds around boundaries of objects from different classes, (2) the diversity of shapes within the same class, and (3) the scale differences between classes. In this study, we propose a novel double self-attention convolutional network called the DAPnet. The double self-attention includes the point attention module (PAM) and the group attention module (GAM). For problem (1), the PAM can effectively assign different weights based on the relevance of point clouds in adjacent areas. Meanwhile, for problem (2), the GAM enhances the correlation between groups, i.e., grouped features within the same classes.
To solve problem (3), we adopt a multiscale radius to construct the groups and concatenate extracted hierarchical features with the output of the corresponding upsampling process. 
Under the ISPRS 3D Semantic Labeling Contest dataset, the DAPnet outperforms the benchmark by 85.2\% with an overall accuracy of 90.7\%. 
By conducting ablation comparisons, we find that the PAM effectively improves the model than the GAM. The incorporation of the double self-attention module has an average of 7\% improvement on the pre-class accuracy. Plus, the DAPnet consumes a similar training time to those without the attention modules for model convergence. 
The DAPnet can assign different weights to features based on the relevance between point clouds and their neighbors, which effectively improves classification performance. The source codes are available at: https://github.com/RayleighChen/point-attention.
\end{abstract}


%
\IEEEpeerreviewmaketitle

\section{Introduction}
Airborne Laser Scanning (ALS) is one of the most important remote sensing technologies that experience fast development~\cite{7907286,8082119,9112288}. ALS point clouds are advantageous over optical data in terms of various lighting conditions and shadows. It has become the most essential form of data at large spatial scales~\cite{8681075,8854321}. However, ALS point clouds contain irregularly distributed points with a series of attributes and have a complex data structure, which makes the point cloud semantic labeling challenging~\cite{wahabzada2015automated,grilli2017review}.



In recent years, deep learning models~\cite{lecun2015deep}, especially convolutional neural networks (CNNs), have been proved to be effective in feature extraction in an end-to-end fashion~\cite{litjens2017survey,deng2014tutorial,zhang2018survey}.
The CNNs also inspire researchers to tackle the challenging point cloud semantic labeling problems. However, traditional CNNs normally consist of 2D convolutional layers, which cannot directly adapt to the structure of 3D point clouds. Hence, 3D CNNs are applied on 2D contexts transformed from 3D point clouds~\cite{li2016vehicle,maturana20153d}, as is seen in VoxNet~\cite{7353481} and ShapeNets~\cite{wu20153d}. Other studies utilize multi-view CNNs to extract geometric features from multiple 2D rendering views of point clouds~\cite{boulch2017unstructured,caltagirone2017fast}. These volumetric CNNs and multi-view CNNs can only be applied to the transformed data from 3D point clouds, which would incur much information loss. The approach to address this issue would be to build a 3D model that can be directly applied to the structure of point clouds. Qi~\cite{qi2017pointnet} proposed a unique deep learning structure called PointNet, which utilizes a set of functions to obtain global features from point clouds. To capture local features and improve the generalizability of the model for pattern recognition tasks, Qi~\cite{qi2017pointnet++} further proposed a novel model called PointNet++. 
More, DGCNN~\cite{wang2019dynamic} can extract local features of point clouds while maintaining permutation invariance. PointCONV~\cite{wu2019pointconv} can learn the weights of point clouds at different locations. KPConv~\cite{thomas2019kpconv} can get a matrix to perform feature updates based on the position of each point, and MS-PCNN~\cite{ma2019multi}, which is less sensitive to data distribution and computational power.
These methods integrate the preprocessing, training, and classification of the point clouds, which have been proven successful in various applications. 

However, compared to the objects from indoor scenes, objects from the ALS point clouds exhibit greater variation even in objects from the same class~\cite{hsiao2004change}. For example, the geometric features of powerlines vary greatly in size and scale. Therefore, previous methods which adopt the extracted geometric features to classify powerlines are prone to misclassifications caused by the large variations of object shapes and scales~\cite{blomley20163d}.
And due to the irregularity of the point cloud distribution, most of the point clouds distributed along the boundaries of objects from different classes are mixed. The features of these point clouds are difficult to be extracted correctly and then used for semantic labeling. Moreover, there are scale differences between ALS point cloud classes; for example, the size and shape of cars do not vary as much as those of trees. But the perceptual field of the feature extraction layer of a CNN is fixed, which poses challenges in classifying objects varying in size and shape. The features captured by the same receptive field may not be able to classify either class well.
Therefore, the features extracted from the ALS point cloud need to be assigned different weights under different classes, and each point can be distinguished from the surrounding points.
The model should pay more attention to the feature extraction process of the points~\cite{nam2017dual}.

This research proposes a novel double self-attention convolutional network called the DAPnet to address the challenging issues laid out above.
The novel model is inspired by the self-attention mechanism~\cite{vaswani2017attention,nam2017dual}, which can produce different values depending on different keys, i.e., the relationship between the input features.
For ALS point clouds, the self-attention module can assign different weights to the features of each point, even if they are close to one another.
To be specific, we have designed a double self-attention module, which refers to the point attention module (PAM) and the group attention module (GAM), as the top feature extraction layer of the model.
The PAM constructs a matrix of key-value pairs based on the correlation between features of points in the same batch.
The extracted features are again assigned different weights under key-value pairs, which can effectively classify the surrounding points, especially those within the border areas. 
As for the shape diversity problem in the same class, considering that sample points in a local region can constitute a local feature, so it's reasonable to divide them into different groups, where each group is a local feature belonging to an object. These different groups contain information about the shape of the object. Therefore, with our proposed GAM, we can redistribute weights among the features via these groups. This mechanism has been proved to be able to improve the final semantic labeling results. 
The two attention modules process the extracted features in parallel, and the model fuses their output.
To obtain multiscale features, during the processing of the data, we also use multiscale radius to construct groups and directly connect the hierarchical features to the upsampling process by the skip-connection method. The
DAPnet can be directly applied to raw ALS point clouds without transformation.
The experimental results on the benchmark dataset show that our method can effectively and efficiently classify the point clouds by generating a state-of-the-art overall accuracy of 90.7\%, which is 5.5\% above the best benchmark.
Compared to the best benchmark methods, DAPnet also significantly improves the average F1 score to 82.3\%, up by 9.1\%. The ablation experiment shows that the double self-attention module improves the average per-class accuracy by 7\% at the same rate of model convergence.
The major contributions of this research can be summarized as follows:
\begin{itemize}
    \item The proposed DAPnet uses the PAM and the GAM to reassign weights to the extracted point cloud features. With these features, the model can effectively classify point clouds from boundaries or at different scales. 
    \item The point attention module is beneficial for classifying areas with mixed point clouds by constructing a matrix of key-value pairs of extracted features and assigning different weights to the inputs.
    \item The group attention module enhances the correlation between groups of the same object by assigning different weights to the groups, which improves the classification accuracy for point clouds with diverse shape classes.
\end{itemize}

The remainder of this paper is organized into three sections. We present our method in Section \ref{methods}. The experimental results are presented in Section \ref{expriment}. We draw conclusions and detailed discussion for further work in Section \ref{conclusion}.

\section{Methods}
\label{methods}

\subsection{Data Preprocessing}

\begin{figure}[tbp]
    \begin{center}
    \includegraphics[scale=0.3]{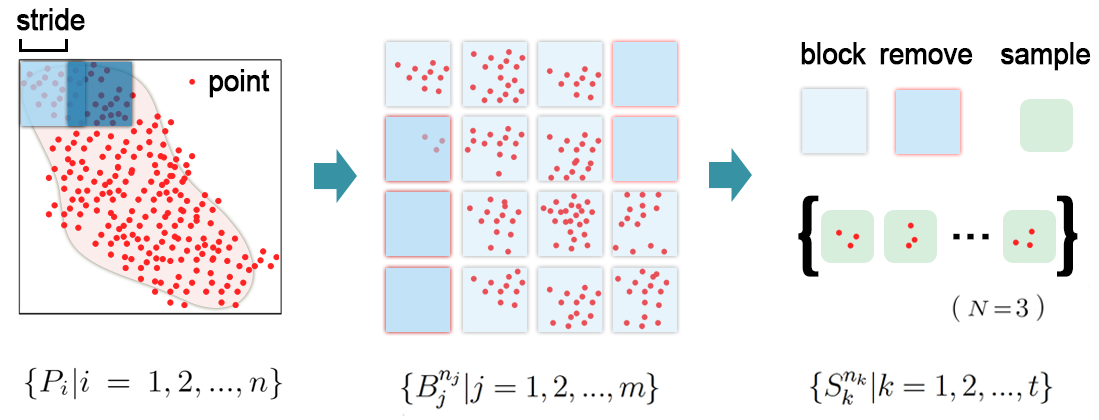}
    \end{center}
    \caption{Demonstration of data processing. From left to right is the process of ALS point clouds in the area to regular input data.}
    \label{fig:proc}
\end{figure}

\begin{figure*}[htp]
    \begin{center}
    \includegraphics[scale=0.8]{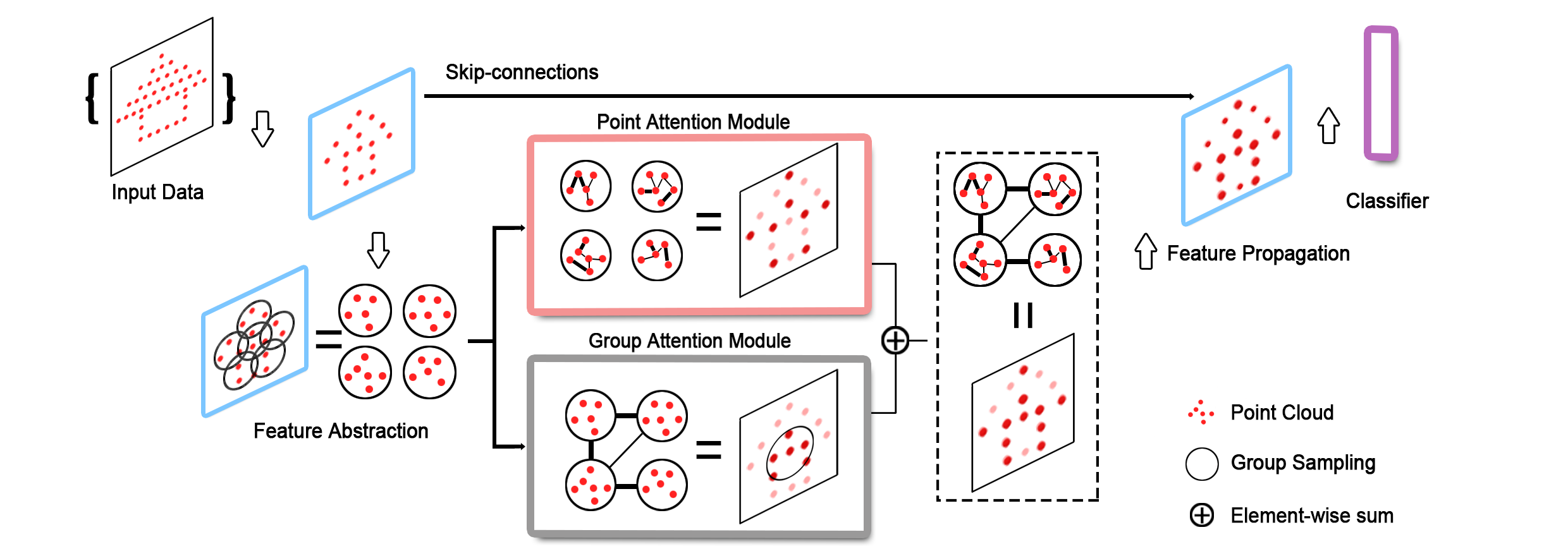}
    \end{center}
    \caption{An overview of the double self-attention convolutional network named DAPnet}
    \label{fig:net}
\end{figure*}

Figure~\ref{fig:proc} shows the workflow of data preprocessing and normalization, which converts point clouds into regular batch input. Given an unordered 3D ALS point cloud $\{P_i | i = 1,2,...,n\} $ with $ P_i \in \mathbb{R}^d$. Each point $P_i$ is a vector $(x, y, z, int, ret, num) $ of its $(x, y, z)$ LiDAR-derived coordinates plus extra channels such as intensity, return number and number of returns, and corresponds to a reference label. For the training set, we obtain the length and width of the area based on the maximum and minimum coordinates. Then, a fixed-size block moves over the entire area, shifting a specified stride at each step.
Each block can overlap with adjacent ones, as the length of the stride could be less or equal to the block size. The entire area depicted by the dataset can be divided into multiple small blocks of the same size with different points.
The processed blocks then reconstitute a new dataset. The new dataset of $m$ blocks is $\{ P \} \rightarrow \{ B_j^{n_j} | j =1,2,...,m\}$, where $B_j^{n_j}$ is the representation of the $j$-th block of points and $n_j$ is the number of points in the corresponding block.
When the number of points in a block is below the threshold, we remove that block. 
The settings of threshold and block size depend on point density and region size. According to our experimental results, a self-attention matrix with more than 250 points can effectively enhance the features of the points. So, when a block has less than 250 points, it is removed.
We apply a similar process to create the testing set without either reserving overlapping areas or removing blocks with insufficient points.

During the model training process, we randomly select blocks and use the min-max normalization method~\cite{jain2011min} to process the coordinates $(x, y, z)$ and intensity based on the point clouds in each block. Then at the block level, a fixed number of points are then randomly sampled. Therefore, for each block, we can obtain $\{ S_k^{n_k} | k =1,2,...,t\}$, where $S$ represents the sampled points, and $n_k$ is the number of samples at time $k$. When the number of point clouds of the block is less than the fixed number, we adopt the Bootstrap sample method~\cite{shao1994bootstrap} to sample enough points. As a result, the final dataset is $\{ P \} \rightarrow \{ S_k^{n_k} | k =1,2,...,t\}$.

\subsection{DAPnet}
\subsubsection{Method Overview}
The proposed DAPnet, a 3D point cloud semantic labeling model, consists of feature abstraction layers~\cite{qi2017pointnet++}, a PAM, a GAM, feature propagation layers, and a classifier as shown in Figure~\ref{fig:net}.

The feature abstraction layer is a feature extractor adapted to raw point clouds. It contains group sampling and convolutional layers, which can effectively extract hierarchical features. Group sampling divides all points into different groups to learn multiple local features, and convolutional layers extract features from the data via multi-size kernels. Compared to PointNet++, we use multiple sizes of groups for sampling while also concatenating all the features, by the skip-connection method, directly with the output of the corresponding feature propagation layer.
After the input data is processed from multiple feature abstraction layers, we can, in turn, obtain multiple features of points and groups as output. 
 
Both as critical parts of the DAPnet, the proposed point and group attention modules are two types of self-attention modules. They assign different weights to the feature based on the relevance of the points or groups to the surrounding features, thus improving the performance of the semantic labeling.  
These two modules process the output from the feature abstraction module in parallel.  
In the PAM, especially for point clouds in the border area of the objects, features assigned different weights can better reflect their association with different classes. Similarly, an object usually corresponds to several groups of points. In the GAM, the weight assigned by the correlation between the groups shows whether they are of the same class. Finally, the enhanced features are the sum of values produced from the two modules. 

Feature propagation is an upsampling operation that can use the learned features to retrieve the features of all input points. The upsampling process also concatenates the features on[from] the corresponding feature abstraction layer through the skip-connection method. Then, the neighboring points have similar features. The final results of feature propagation are sent to the classifier.

The final classifier realizes the classification of each point class, thereby achieving the semantic labeling of the ALS point clouds.

\subsubsection{Feature Abstraction Layer}
As the first part of the DAPnet, the feature abstraction layers aggregate points into groups at different hierarchies and progressively extract features at different layers. 

For a given set of points $P = \{ p_1, p_2, …, p_N\}$, where $P$ is an $N \times D$ matrix. $N$ is the number of points, and $D$ is the number of dimensions of a point feature, such as coordinate, RGB, intensity, etc. For group sampling, in order to continuously measure the spatial distance at different layers, we transform the ALS point clouds $P$ to a $N \times (l+D)$ matrix, where $l$ represents the dimension of coordinates. And the spatial coordinate dimensions are not involved in the process of feature extraction.
We define $M=l+D$, so ALS point clouds can be $P^{N \times M}$.
In our study data, the $l=3$ and $D =6$. Thus, we denote the input points as $P^{N \times 9}$. Then, to make sure that the sampled group covers the entire object, we use iterative farthest point sampling (FPS)~\cite{eldar1997farthest} to find several major centroids, which means that during the iteration, each selected point is the farthest one from the rest. In this study, we used the $L_2$ distance to measure the spatial distance between two points. The sampled centroids are a subset of $P$, denoted as $\{p_{i_1}, p_{i_2}, …, p_{i_m} \}$. $i_m$ represents the index in $P$. Based on these centroids, we construct groups based on the ball query method~\cite{qi2017pointnet++} which can find all points within a fixed radius. In this step, we use the multiscale radius for grouping, taking into account the problem of sparse and dense point sampling. After group sampling, the input data become $P^{G\times S \times M}$, where $G$ represents the number of groups, and $S$ represents the fixed number of points in each group.

To extract features, we apply an 1-D convolutional operation on each group which is treated as a new object. The convolution results are normalized with Batch Normalization (BN)~\cite{ioffe2015batch} and put through the ReLU activation function~\cite{krizhevsky2012imagenet}, a non-linear operation. This process can be written mathematically as follows,
\begin{equation}
    f(\{p_1, p_2, ..., p_j\}) =  \text{ReLU}(\text{BN}_{k=1}^j( W \circ \{p_k\} + b)),
\end{equation}
where $\{p_1, p_2, ..., p_j\}$ is a group of points, and $\circ$ represents the 1-D convolutional operation. Next, we apply max pooling~\cite{murray2014generalized} over the output from the feature extraction, to extract global features. 
Since each group can be treated as a part of an object, the extracted grouped features are the different local features of the object. The entire feature abstraction layer produces the output data, $P^{G \times S \times T}$, where $T$ represents the dimension of the features.

\begin{figure}[tp]
  \centering
  \subfigure[Point Attention Module]{
    \label{fig:pam} 
    \includegraphics[scale=0.3]{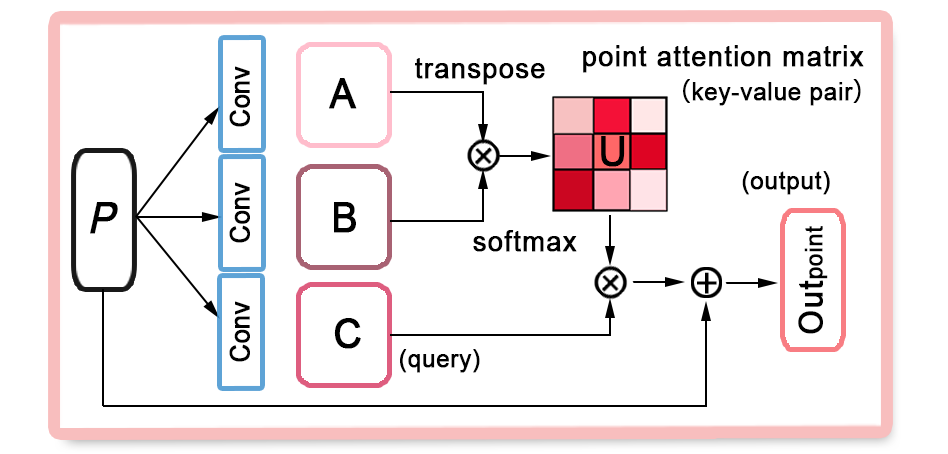}}
  \subfigure[Group Attention Module]{
    \label{fig:gam}
    \includegraphics[scale=0.3]{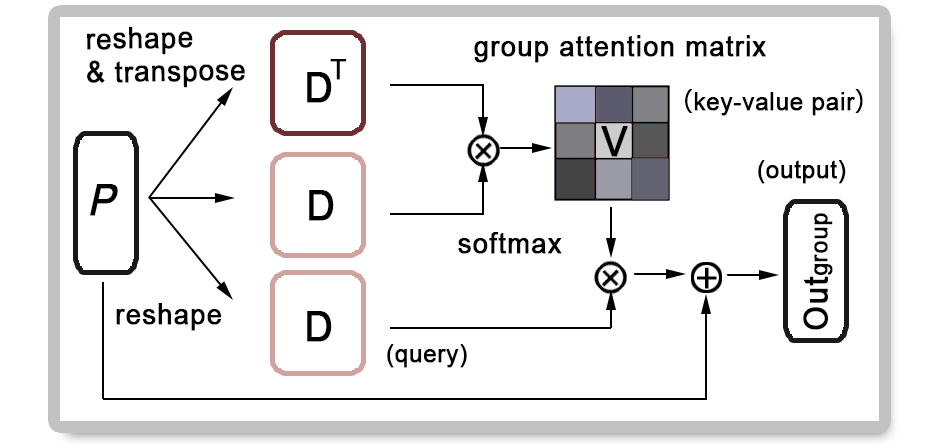}}
  \caption{The details of the double self-attention module.}
  \label{fig:attention} 
\end{figure}

\subsubsection{Point Attention Module}
The point clouds in the border area of the objects from different classes are difficult to classify because the points from two sides of the boundary may be mingled up. Therefore the features of these points need to have different weights for different classes. In the light of this idea, we design the PAM based on the self-attention mechanism~\cite{vaswani2017attention}. This mechanism can be briefly described as mapping queries to key-value pairs. Through the query in the key-value pair, the weight of the query under the corresponding key is obtained. Then we assign the weights to the corresponding query to get the output. The same query has different outputs under different keys. The PAM constructs key-value pairs based on the correlation between the features of adjacent point clouds. When the feature is input to the PAM, the outputs obtained by the key-value pair mapping are beneficial to model classification.

Taking the output $P^{G \times S \times T }$ from the feature abstraction as input, the process of the PAM is illustrated in Figure~\ref{fig:pam}. First, it constructs key-value pairs, which are derived from the input data under the self-attention mechanism. 
We feed the output data into two convolutional layers respectively and achieve two outputs $A^{G \times S \times T}$ and $B^{G \times S \times T}$. Then the point features of each group are expanded into vectors. We reshape them to two matrices, $\mathbf{A}^{G \times J}$ and $\mathbf{B}^{G \times J}$, where $J$ is $S \times T$, and further transpose the matrix $A$. After the matrix multiplication $\mathbf{A}^\top \mathbf{B}$, we apply a softmax function~\cite{bouchard2011clustering} on the results to obtain the point attention matrix $\mathbf{U}^{J \times J}$ which is the key-value pairs as follows, 
\begin{equation}
    \mathbf{U}_{j,i} =  \frac{ \text{exp} (s_{ij}) }{ \sum^J_{i=1} \sum^J_{j=1} \text{exp} (s_{ij})},
\end{equation}
where $s_{ij} \in \mathbf{A}^\top \mathbf{B}$. The $\mathbf{U}_{j,i}$ indicates that the $i^{th}$ feature impacts on $j^{th}$ feature.
A larger value between the two points indicates that they have a strong correlation. We also feed the previous output $P$ through a convolutional layer and achieve the output data $C^{G \times S \times T}$ to query. 
We convert it to the matrix $\mathbf{C}^{G \times J}$ and multiply by the transpose of the point-attention matrix. The result is $\mathbf{C} \mathbf{U}^{\top} \in \mathbb{R}^{G \times J}$. 
Then we reshape the result to $G \times S \times T$ and multiply a scale $\alpha$ to do an element-wise sum with the data $P$. The final result is the output of attention mechanism, which can be written mathematically as follows, 
\begin{equation}
    \text{Out}_{\text{point}} =  \alpha \sum^J_{i=1}  \mathbf{U}_{j,i} \mathbf{C}_i + D_j,
\end{equation}
where $\alpha $ is the learnable scale parameter and initialized as 0. It can gradually assign more weight to the non-local features beyond the local neighborhood. The points have different weights, which is beneficial for classifying some mixed points in the border area. The final result $\text{Out}_{\text{point}}$ is a feature-enhanced output with the point attention matrix.

\begin{table*}[h]
\centering
\caption{The detailed operations of DAPnet}
\begin{tabular}{lccccl}
\hline
\multicolumn{1}{c}{Layer Name} & Input & Output & Operations & \# Kernels & \multicolumn{1}{c}{Note} \\ \hline
\multirow{4}{*}{Feature Abstraction} & $L_0$ & $L_1$ & Conv,Conv,Conv,Max & (32,32,64) &  \\
 & $L_1$ & $L_2$ & Conv,Conv,Conv,Max & (64,64,128) &  \\
 & $L_2$ & $L_3$ & Conv,Conv,Conv,Max & (128,128,256) &  \\
 & $L_3$ & $L_4$ & Conv,Conv,Conv,Max & (256,256,512) &  \\ \hline
\multirow{5}{*}{PAM} & $L_4$ & $L_5$ & Conv & 64 &  \\
 & $L_4$ & $L_6$ & Conv & 64 &  \\
 & $L_5$,$L_6$ & $L_7$ & Multiple,Softmax & - & key-value pairs \\
 & $L_4$ & $L_8$ & Conv & - & query \\
 & $L_4$,$L_7$,$L_8$ & $L_9$ & Multiple,Add & - & $\alpha$,output \\ \hline
\multirow{2}{*}{GAM} & $L_4$ & $L_{10}$ & Multiple,Softmax &  & key-value pairs \\
 & $L_4$,$L_{10}$ & \multicolumn{1}{l}{$L_{11}$} & Multiple,Sum &  & $\beta$,query,output \\ \hline
\multirow{4}{*}{Feature Propagation} & \begin{tabular}[c]{@{}c@{}}$L_3$,\\ $L_{9}$,$L_{11}$\end{tabular} & \multicolumn{1}{l}{$L_{12}$} & \begin{tabular}[c]{@{}c@{}}Sum,Concat,Interpolation,\\ Conv,Conv\end{tabular} & (256,256) &  \\
 & $L_2$,$L_{12}$ & \multicolumn{1}{l}{$L_{13}$} & \begin{tabular}[c]{@{}c@{}}Concat,Interpolation,\\ Conv,Conv\end{tabular} & (256,256) &  \\
 & $L_1$,$L_{13}$ & \multicolumn{1}{l}{$L_{14}$} & \begin{tabular}[c]{@{}c@{}}Concat,Interpolation,\\ Conv,Conv\end{tabular} & (256,128) &  \\
 & $L_{14}$ & \multicolumn{1}{l}{$L_{15}$} & Conv,Conv,Conv & (128,128,128) &  \\ \hline
\multirow{2}{*}{Classifier} & $L_{15}$ & \multicolumn{1}{l}{$\hat{c}$} & Conv & (128) &  \\
 & $\hat{c}$,$c$ & \multicolumn{1}{l}{$L_{16}$} & Softmax,Cross entropy &  &  \\ \hline
\end{tabular}
\label{tab:detail}
\end{table*}

\subsubsection{Group Attention Module}
Each group is sampled from the points around the centroid, and they almost cover the entire object. And groups from the same object have higher correlations, and each group contributes differently to the final semantic labeling result. To enhance the contribution of different groups, we designed the group focus [attention] module based on the relationship between groups.
Similarly, it also needs to construct queries, key-value pairs, and outputs. Different from the PAM, it does not require a convolution operation at the beginning since it can destroy the relationship between groups. 

For the output $P$ of the feature abstraction, the process of the GAM is illustrated in Figure~\ref{fig:gam}. We directly reshape it to a $G \times J$ matrix $\mathbf{D}$. Then we perform a matrix multiplication between the matrix and the transpose matrix and a softmax function on the result to obtain the group attention matrix $V^{G \times G}$, whose elements are calculated with the following,
\begin{equation}
    \mathbf{V}_{j,i} =  \frac{ \text{exp}   ((\mathbf{D}\mathbf{D}^\top)_{ij}) }{ \sum^G_i \sum^G_j \text{exp} ((\mathbf{D}\mathbf{D}^\top)_{ij})}.
\end{equation}

$\mathbf{V}_{j,i}$ is the key-value pair for the GAM. It represents the relationship between groups. Then the reshaped output $D^{G \times J}$ multiple V with a scale $\beta$. Finally, we do an element-wise sum with the data $P$. The process can be written as follows:
\begin{equation}
    \text{Out}_{\text{group}} =  \beta \sum^G_{i=1}   \mathbf(V)_{j,i} \mathbf{D}_i + D_j,
\end{equation}
where $\beta$ is also initialized as 0. When each group adds the corresponding weights, the features can be boosted based on the relationship between groups. And the weights assigned according to the correlation between the groups are beneficial for objects with irregular shapes.

\subsubsection{Feature Propagation Layer}
Because the final feature-enhanced outputs from the point and group attention modules cannot be directly used to classify each point, the feature propagation layer generates features from the feature-enhanced outputs to each point.

Feature propagation is also a hierarchical process, which progressively generates features corresponding to all points layer by layer. First, we concatenate the summed outputs and the features of the previous feature abstraction layer through the skip-connection method to the input. Then we use the inverse distance weighted method~\cite{setianto2013comparison} based on k nearest neighbors (KNN)~\cite{fukunaga1975branch} to interpolate feature values of the points in each layer, as follows, 

\begin{equation}
\label{eq:fp}
u(x) = \left\{ \begin{array}{ll}
\frac{\sum^N_{i=1} w_i (x) u_i}{\sum^N_{i=1}w_i(x)}, & \textrm{if $ w_i = \frac{1}{||x - x_i||^2} $ for all $i$}\\
u_i, & \textrm{if $w_i = 0$ \qquad for some $i$}
\end{array} \right.
\end{equation}

From Eq.~\ref{eq:fp}, we can find that points farther from the point $x$ have less weight. After the weight of each point is assigned, we perform a global normalization for all weights. Through the processing of all feature propagation layers, we can get each point score $\hat{c}$ and compare it with the corresponding label $c$ under the cross-entropy loss function~\cite{bosman2000negative}. Finally, we use the gradient descent algorithm~\cite{kingma2014adam} to update all model weights.

\subsection{Training parameters}
In this section, we introduce the architecture of the DAPnet. The details of the architecture are provided in Table~\ref{tab:detail}, including data flow, the main process of operation, and the number of convolutional kernels.

First, the DAPnet has 4 layers of feature abstraction. In each layer, the point clouds are divided by group sampling. We use 3 convolutional layers for feature extraction on these groups, and the number of convolution kernels is gradually increased. 
After 3 convolutional layers, we apply max-pooling to obtain the global features of each group $L_1$. Then, for the final output $L_4$ of feature abstraction, we can get the enhanced feature results of $L_9$ and $L_{11}$ by the point and group attention modules. The following 4 layers are feature propagation. Except for the last layer, the input is concatenated to the output of the corresponding feature abstraction. Finally, by comparing the obtained score of all original points with the corresponding label $c$, the classifier performs classification.

\subsection{Benchmark Methods}
To compare benchmark methods on the ISPRS 3D Semantic Labeling Contest, we selected the following algorithms for a brief review based on performance, feature extraction methods, and other factors, with their abbreviations as method names. 

\begin{table}[h]
\centering
\caption{The detail of the benchmark methods}
\label{tab:method}
\setlength{\tabcolsep}{0.5mm}{
\begin{tabular}{lccc}
\hline
Method & Transformation & Features & Classifier \\
\hline
IIS\_7 & Yes & Geometrical & $ - $ \\
UM & Yes & Geometrical & OvO classifier \\
HM\_1 & Yes  & Multiple Scales & RF \\
WhuY3 & Yes & Geometrical & Softmax \\
LUH & Yes & Multiple Scales & RF \\
RIT\_1 &  No  & Geometrical & Softmax \\
NANJ2 & Yes & Geometrical & Softmax \\
PointNet & No & Geometrical & Softmax \\
PointNet++ & No & Multiple Scales & Softmax \\
PointSIFT & No & Multiple Scales & Softmax \\
GAPHNet & No & Multiple Scales & Softmax \\
\hline
DAPnet &  No  & Multiple Scales & Softmax \\
\hline 
\end{tabular}}
\end{table}

\begin{figure*}[!htbp]
    \begin{center}
    \includegraphics[scale=0.4]{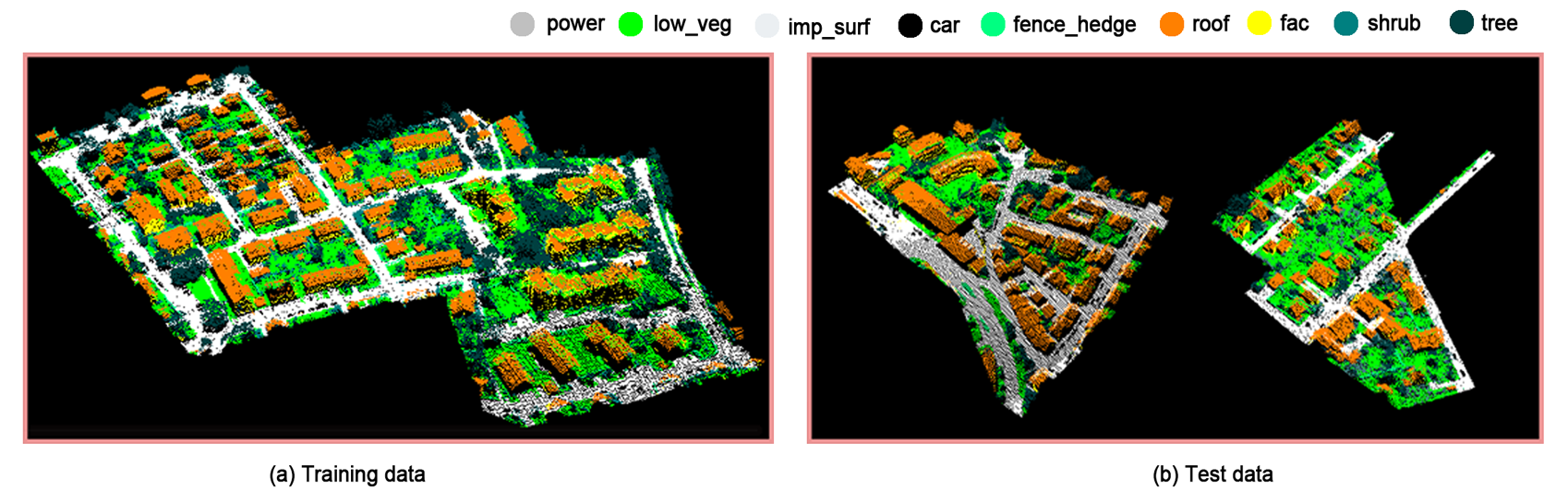}
    \end{center}
    \caption{3D point cloud acquired via airborne laser scanning.}
    \label{fig:data}
\end{figure*}

\begin{table*}[!htbp]
\centering
\caption{Number of 3D points per class.}
\label{tab:data}
\begin{tabular}{lcccccccccc}
\hline
Class        & power & low\_veg & imp\_surf & car   & fence\_hedge & roof    & fac    & shrub  & tree    & Total   \\ \hline
Training Set & 546   & 180,850  & 193,723   & 4,614 & 12,070       & 152,045 & 27,250 & 47,605 & 135,173 & 753,876 \\
Test Set     & 600   & 98,690   & 101,986   & 3,708 & 7,422        & 109,048 & 11,224 & 24,818 & 54,226  & 411,722 \\ \hline
\end{tabular}
\end{table*}

The differences between these benchmark methods mainly lie in three aspects, namely, data transformation, features, and classifier types. Table~\ref{tab:method} show the characteristics of the mentioned methods and our proposed method. The \textbf{IIS\_7}~\cite{2016A} method over-segments the LiDAR data into super voxels in terms of various attributes (i.e., shape, colors, intensity, etc.), and applies the spectral and geometrical feature extraction. The \textbf{UM}~\cite{2016Context} method combines various features including point-attributes, textural patterns, and geometric attributes to a one-vs-one classifier. The \textbf{HM\_1}~\cite{2017Semantische} method depends on the geometric features of a selection of neighborhoods. To conduct the classification, it adopts a Conditional Random Field (CRF)~\cite{finkel2008efficient} with a RF classifier. The \textbf{WhuY3}~\cite{yang2017convolutional} method transforms the 3D neighborhood features of point clouds to a 2D image and applies a CNN to extract the high-level representation of features. The \textbf{LUH}~\cite{2016HIERARCHICAL} method extends the Voxel Cloud Connectivity Segmentation~\cite{papon2013voxel}. It designs a two-layer hierarchical CRF framework to connect relationships, along with the Fast Point Feature Histograms (FPFH) features~\cite{rusu2009fast}. The \textbf{RIT\_1}~\cite{yousefhussien2018multi} method proposes a 1D-fully convolutional network extended from PointNet~\cite{qi2017pointnet} in an end-to-end fashion. The \textbf{NANJ2}~\cite{2018Classifying} method applies a multiscale CNN to learn the geometric features on a set of multiscale contextual images. 
Furthermore, we also use PointNet, PointNet++~\cite{qi2017pointnet++}, PointSIFT~\cite{2018PointSIFT}, and GAPHNet~\cite{LI202026} for comparison. 
These methods are based on CNNs and can be applied to the original point cloud data without transformation. They use the extracted features to achieve the classification of points through the softmax classifier. PointSIFT obtains scale invariance by encoding each point in 8 directions. GAPHNet also uses the attention mechanism to improve the performance of semantic labeling.
During the training process, we used the same data processing methods for PointNet and PointNet++.


\begin{figure*}[!htbp]
    \begin{center}
    \includegraphics[scale=0.83]{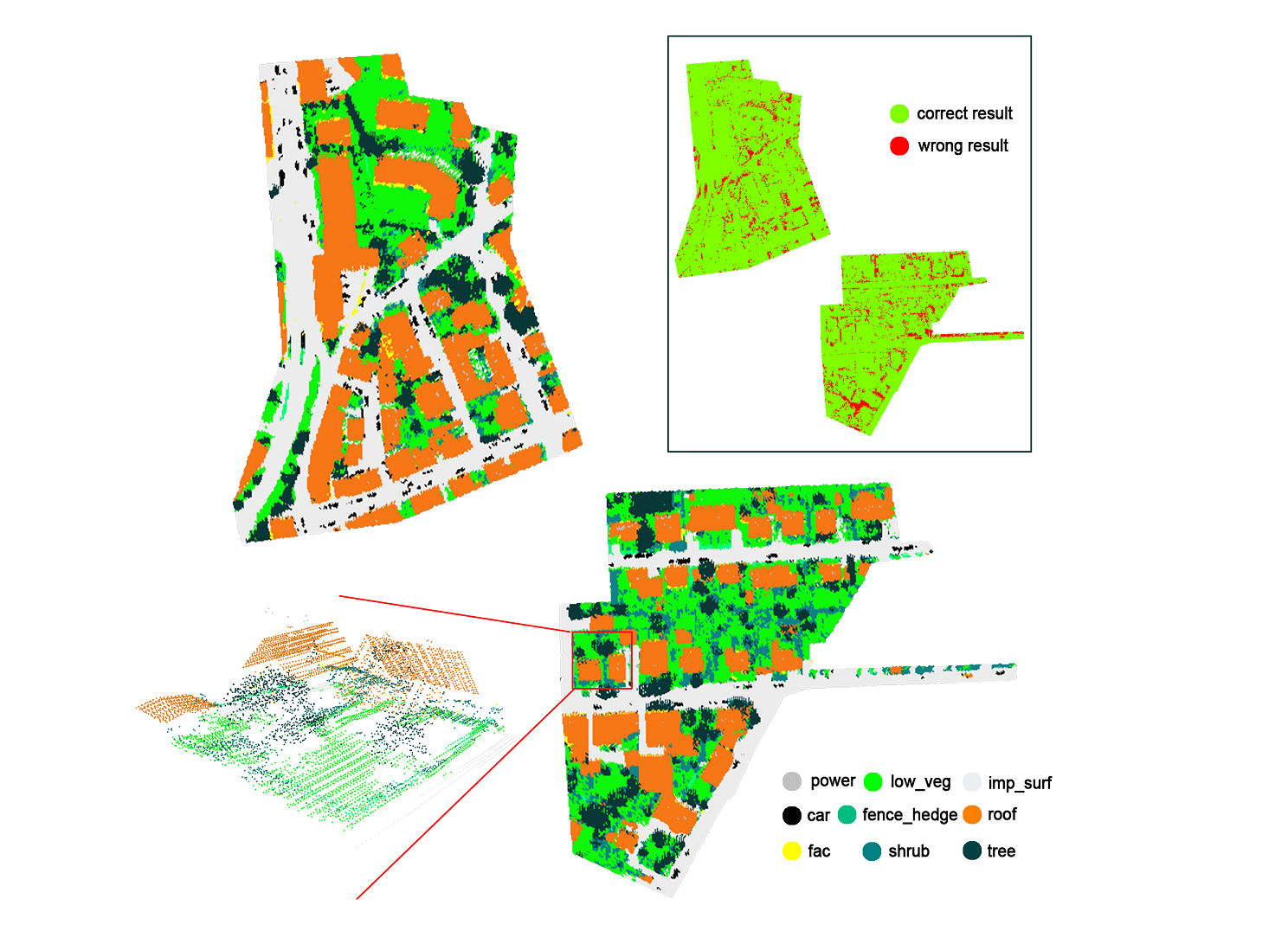}
    \end{center}
    \caption{The classification result and error map of DAPnet on the Vaihingen 3D test dataset. The block in the red box is enlarged to show the details}
    \label{fig:bench}
\end{figure*}

\begin{table*}[h]
\centering
\caption{The confusion matrix of DAPnet on the Vaihingen 3D test dataset with per-class precision, recall and F1 score (\%). The overall accuracy (OA) is 90.7\%.}
\setlength{\tabcolsep}{1.7mm}{
\begin{tabular}{lccccccccc}
\hline
Classes & power & low\_veg & imp\_surf & car & fence\_hedge & roof & fac & shrub & tree \\ \hline
power & \textbf{90.3} (542) & 0.2 (1) & 0.0 (0) & 0.0 (0) & 0.0 (0) & 4.7 (28) & 0.3 (2) & 0.3 (2) & 4.2 (25) \\
low\_veg & 0.0 (0) & \textbf{89.5} (88,357) & 6.6 (6,470) & 0.1 (73) & 0.1 (106) & 0.5 (452) & 0.1 (104) & 2.8 (2,767) & 0.4 (361) \\
imp\_surf & 0.0 (0) & 2.2 (2,279) & \textbf{97.3} (99,225) & 0.1 (114) & 0.0 (48) & 0.1 (126) & 0.0 (27) & 0.1 (145) & 0.0 (22) \\
car & 0.0 (0) & 1.9 (70) & 1.1 (40) & \textbf{90.2} (3,344) & 0.8 (28) & 1.6 (58) & 0.5 (20) & 3.7 (138) & 0.3 (10) \\
fence\_hedge & 0.0 (0) & 9.2 (683) & 1.8 (136) & 1.6 (122) & \textbf{41.1} (3,047) & 1.8 (132) & 1.2 (91) & 28.4 (2,109) & 14.8 (1,102) \\
roof & 0.1 (75) & 0.3 (361) & 0.1 (97) & 0.0 (5) & 0.0 (52) & \textbf{97.0} (105,750) & 0.4 (414) & 1.0 (1,045) & 1.1 (1,249) \\
fac & 0.1 (10) & 8.1 (912) & 0.9 (99) & 1.1 (121) & 0.3 (33) & 13.8 (1,548) & \textbf{62.0} (6,963) & 7.0 (787) & 6.7 (751) \\
shrub & 0.0 (1) & 11.6 (2,882) & 0.6 (141) & 0.5 (132) & 1.2 (309) & 2.6 (653) & 1.1 (275) & \textbf{74.0} (18,353) & 8.3 (2,072) \\
tree & 0.0 (10) & 2.2 (1,174) & 0.0 (16) & 0.1 (35) & 0.3 (142) & 1.6 (882) & 0.7 (356) & 6.9 (3,761) & \textbf{88.2} (47,850) \\ \hline
Precision & 85.0 & 91.4 & 93.4 & 84.7 & 80.9 & 96.5 & 84.4 & 63.1 & 89.5 \\
Recall & 90.3 & 89.5 & 97.3 & 90.2 & 41.1 & 97.0 & 62.0 & 74.0 & 88.2 \\
F1 score & 87.6 & 90.4 & 95.3 & 87.4 & 54.5 & 96.7 & 71.5 & 68.1 & 88.9 \\ \hline
\end{tabular}}
\label{tab:oa2}
\end{table*}

\begin{table*}[!h]
\centering
\caption{The comparison between DAPnet and other methods on Vaihingen 3D test dataset. The per-class F1 score, the average F1 scores (Avg. F1), and the overall accuracy (OA). (\%)}
\setlength{\tabcolsep}{3mm}{
\begin{tabular}{lccccccccccc}
\hline
Method & power & low\_veg & imp\_surf & car & fence\_hedge & roof & fac & shrub & tree & Avg. F1 & OA\\ \hline
IIS\_7 & 54.4 & 65.2 & 85.0 & 57.9 & 28.9 & 90.9 & — & 39.5 & 75.6 & 55.3 & 76.2\\
UM & 46.1 & 79.0 & 89.1 & 47.7 & 5.2 & 92.0 & 52.7 & 40.9 & 77.9 & 58.9 & 80.8\\
HM\_1 & 69.8 & 73.8 & 91.5 & 58.2 & 29.9 & 91.6 & 54.7 & 47.8 & 80.2 & 66.4 & 80.5\\
WhuY3 & 37.1 & 81.4 & 90.1 & 63.4 & 23.9 & 93.4 & 47.5 & 39.9 & 78.0 & 61.6 & 82.3\\
LUH & 59.6 & 77.5 & 91.1 & 73.1 & 34.0 & 94.2 & \textbf{56.3} & 46.6 & \textbf{83.1} & 68.4 & 81.6\\
RIT\_1 & 37.5 & 77.9 & 91.5 & 73.4 & 18.0 & 94.0 & 49.3 & 45.9 & 82.5 & 63.3 & 81.6\\
NANJ2 & 62.0 & \textbf{88.8} & 91.2 & 66.7 & \textbf{40.7} & 93.6 & 42.6 & 55.9 & 82.6 & 69.3 & \textbf{85.2} \\
PointNet & 43.6 & 80.4 & 87.8 & 47.3 & 21.6 & 81.9 & 28.7 & 38.5 & 64.5 & 54.9 & 75.1\\
PointNet++ & 71.4 & 84.9 & 91.6 & 73.9 & 36.2 & 90.4 & 51.9 & 50.5 & 75.9 & 69.6 & 84.2\\ 
PointSIFT & 73.3 & 77.4 & 91.7 & 70.4 & 33.8 & 93.7 & 51.1 & 53.6 & 80.6 & 68.8 & 82.7 \\
GAPHNet& \textbf{77.5} & 78.9 & \textbf{94.0} & \textbf{74.2} & 35.3 & \textbf{95.0} & 52.3 & \textbf{57.6} & 82.1 & \textbf{73.2} & 84.5 \\\hline
DAPnet & \textbf{87.6} & \textbf{90.4} & \textbf{95.3} & \textbf{87.4} & \textbf{54.5} & \textbf{96.7} & \textbf{71.5} & \textbf{68.1} & \textbf{88.9} & \textbf{82.3} & 90.7\\ \hline
\end{tabular}}
\label{tab:ben_f1}
\end{table*}

\section{Experiment}
\label{expriment}
\subsection{Dataset}
The Vaihingen dataset has been proposed within the scope of the ISPRS test project on urban classification ~\cite{cramer2010dgpf,niemeyer2014contextual}, and it is the benchmark dataset for ISPRS 3D semantic labeling.
More details about this dataset could be found on the ISPRS website\footnote{http://www2.isprs.org/commissions/comm3/wg4/3d-semantic-labeling.html}.
In detail, the ALS point cloud dataset consists of 1,165,598 points and is divided into two areas for training and testing. There are a total of 753,876 training points and 411,722 test points. The training area is mainly residential, with detached houses and high-rise buildings. The test area is located in the center of Vaihingen and has dense and complex buildings. We aim to discern the following 9 object classes with the model, including Powerline (power), Low Vegetation (low\_veg), Impervious Surfaces (im\_surf), Car, Fence/Hedge (fence\_hedge), Roof, Facade (fac), Shrub and Tree. Each point in the point clouds contains LiDAR-derived (x, y, z) coordinates, backscattered intensity, return number, number of returns, and reference label. The training and test areas are demonstrated in Figure~\ref{fig:data}, and the class distribution of the training and test sets are shown in Table~\ref{tab:data}.

We also use the dataset of the 2019 IEEE GRSS Data Fusion Contest 3D point cloud classification challenge~\footnote{https://competitions.codalab.org/competitions/20217} (DFC 3D) to validate the generalization ability of the DAPnet. This dataset is also an ALS point cloud dataset, captured in Jacksonville, Florida, and Omaha, Nebraska. 
The DFC 3D dataset has 6 classes, which are Ground, High Vegetation/Trees (high\_veg), Building Roof (building), Water, Elevated Road/Bridge, and Unlabeled.
Every point is also a vector $(x, y, z, int, ret, num) $.

\subsection{Implementation Details}
We implemented the DAPnet with the PyTorch framework. To process the data, We choose 30$\times$30m as the block size and 10 as the stride. In each block, we sample 1024 points so that those blocks contain the same number of points. The number of groups for the group sampling of the feature abstraction layer are 256, 128, 64, and 32, which correspond to the sampling radius of 0.1, 0.2, 0.4, and 0.8. The multi-scale sampling radii are [0.05, 0.1], [0.1, 0.2], [0.2, 0.4], and [0.4, 0.8] respectively. A batch size of 16 and the maximum epoch of 200 are set for model training. The Adam optimizer~\cite{kingma2014adam} with a momentum of 0.9 and a decaying rate of 0.0001 is used. In terms of the learning rate policy, we use polynomial decay, where the learning rate is monotonically decreasing as the model training progress by iterations. The learning rate at the $i$-th iteration in polynomial decay is given by Equation~\ref{eq:lr},
\begin{equation}
    lr_{i} = (lr_{1}-lr_{I}) \times (1 - \frac{i}{I})^{0.7} + lr_{terminal}
    \label{eq:lr}
\end{equation}
where the initial learning rate $lr_{1}$ is set to 0.001, and the learning rate $lr_{I}$ at the last iteration $I$ is set to $1e^{-5}$ for model training. 
The evaluation metrics of per-class accuracy, precision, recall, and F1-score are calculated using the ISPRS 3D Semantic Labeling Contest dataset. The source codes are available at Github\footnote{https://github.com/RayleighChen/point-attention}.

\subsection{Classification Results}

Table~\ref{tab:oa2} shows the confusion matrix of the DAPnet model. The overall accuracy is 90.7\%, and the model achieves over 90\% accuracy on classes of powerline, impervious surface, car, and roof. 
Objects from these classes usually have fixed shapes, and the DAPnet can process the ALS point clouds of these well.
However, the model has not been effective in terms of accuracy in classifying fence/hedge (41.1\%), facade (62.0\%), and shrub (74.0\%). Ineffectiveness in the worst-performing class, fence/hedge, can be attributed to diverse shapes, plus those point clouds tend to be mixed with objects from other classes. 
Objects wrongfully classified as fence/hedge are mainly in shrub and tree classes, at 28.4\% and 14.8\%. Shrubs and trees bear great similarity to objects in fence/hedge in terms of height, topological features, and spectral reflectance, and the precision for classifying them is high at 84.7\%, indicating that the DAPnet is less likely to wrongly classify shrub and tree as the fence/hedge. 
On the other hand, many facade points are misclassified as roofs because of the similarity between the two classes, while roof points are rarely misclassified by DAPnet as the facade class.
The result reveals that DAPnet has excellent performance for classes with fixed shapes and also shows good precision for classes with diverse shapes.

The classification results and error maps of the DAPnet are shown in Figure~\ref{fig:bench}.
The DAPnet can correctly classify ALS point clouds in most areas, especially where the impervious surface and roof are mixed. Nonetheless, the classification error in areas of mixed shrubs and trees is large according to the error map. When point clouds have similar features, and their constituent objects have irregular shapes and intermixed points at the same time, DAPnet is unable to classify these points correctly.
In the classification map, we zoomed in on the semantic labeling results for an area. This area contains impervious surfaces, roofs, low vegetation, and trees. According to the corresponding error map, the DAPnet has only misclassified a small number of low vegetation and tree points. Still, the DAPnet has demonstrated itself to do well in classifying impervious surfaces, roofs, and most of the low vegetation and trees. The power of the DAPnet has been manifest in those promising results.

\subsection{Performance Evaluation}
Table~\ref{tab:ben_f1} shows the F1 score of our method and the benchmark methods. The DAPnet produces a 9.1\% higher average F1 scores than the best benchmark method.
The DAPnet attains the best overall accuracy (90.7\%), which is 5.5\% higher than the best benchmark. While the edge of the overall accuracy is high, the DAPnet also prevails at per class F1 score with an average improvement of 8.1\% compared to the best benchmark of each class.
For powerline, car, fence/hence, facade, and shrub classes, the improvement is even more significant than the best benchmark, at an average percentage of 12.6\%, where the facade class has improved the most 15.2\%. The classes of powerline, car and fence/hedge have a limited number of training samples according to Table~\ref{tab:data}, but the significant increase in F1 scores with the DAPnet shows that the model performs well even with a small number of samples.
With this considerable improvement, the DAPnet achieves a new start-of-the-art performance.

Compared to algorithms that use multiscale features (HM\_1, LUH, PointNet++, PointSIFT, and GAPHNet), DAPnet achieves the best performance on classes at different scales. The incorporation of point and group attention modules allows these multiscale features to contribute better to the results of semantic labeling under the corresponding key-value pair matrix. While both opt for attention mechanism, the DAPnet still maintains an edge over the GAPHNet. In classes such as car, fence/hence, and facade, the per-class F1 score of the DAPnet performs at least 13\% higher than the GAPHNet. The attention module constructed upon the correlation between points and groups can be better used for semantic labeling of point clouds.


\begin{table}[!htbp]
\centering
\caption{The comparison between DAPnet and other methods on DFC 3D test dataset. (\%)}
\begin{tabular}{lccccc}
\hline
Model & ground & high\_veg & building & others & Avg. F1 \\ \hline
PointNet & 14.2 & 31.7 & 13.3 & 11.4 & 19.7 \\
PointNet++ & 15.5 & 45.7 & 28.2 & 23.1 & 29.8 \\
PointSIFT & 14.1 & 46.9 & 29.8 & 15.2 & 30.3 \\
GAPHNet & 20.8 & 76.5 & 35.7 & 25.2 & 44.3 \\
DAPnet & \textbf{21.6} & \textbf{78.3} & \textbf{40.3} & \textbf{26.6} & \textbf{46.7} \\ \hline
\end{tabular}
\label{tab:transfer}
\end{table}

\begin{figure*}[!htbp]
    \begin{center}
    \includegraphics[scale=0.83]{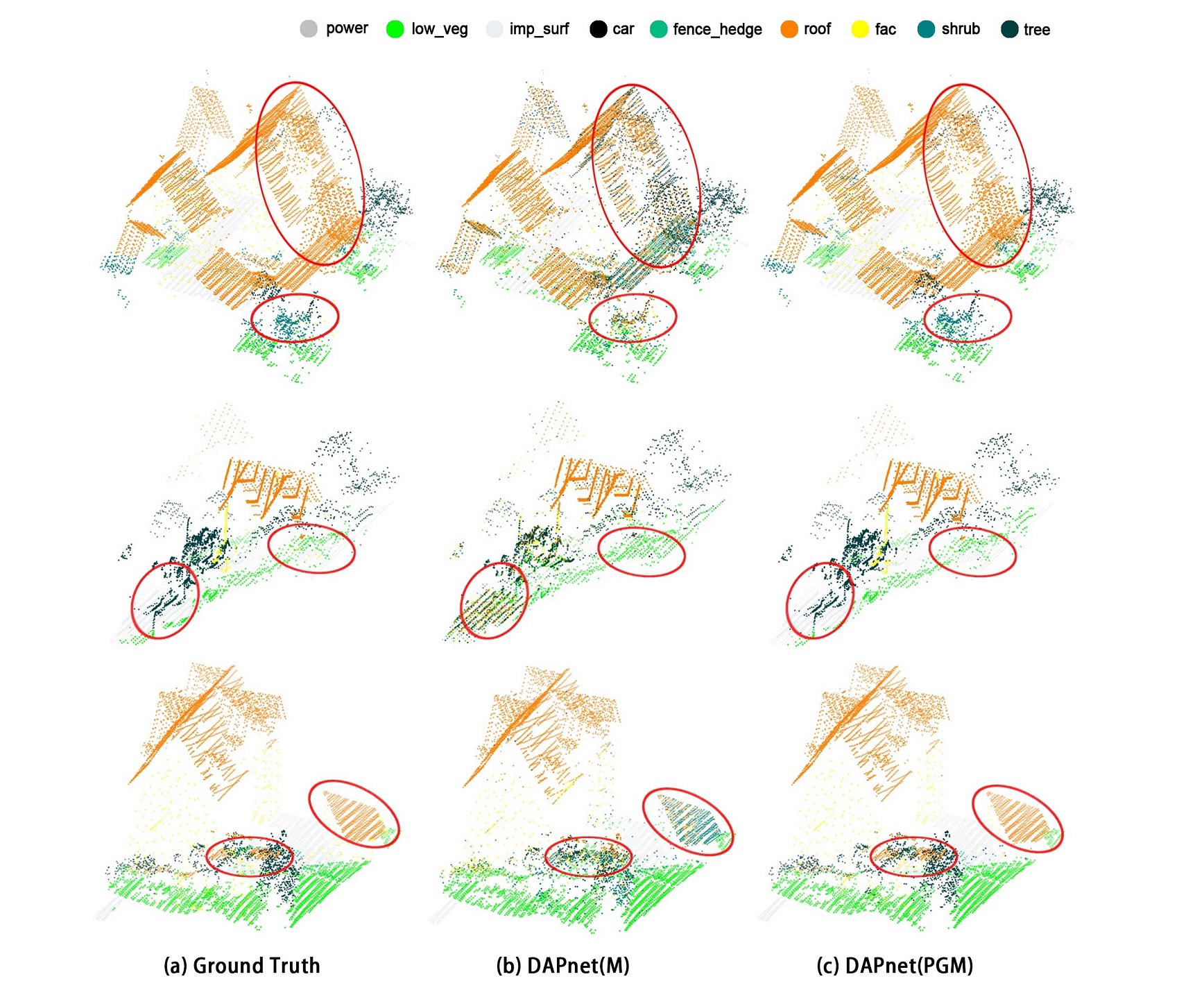}
    \end{center}
    \caption{The qualitative classification results of different models. Circled parts are areas where the results disagree significantly with one another.}
    \label{fig:diff}
\end{figure*}

\begin{table*}[!h]
\centering
\caption{The comparison of performances using different strategies. (\%)}
\setlength{\tabcolsep}{3mm}{
\begin{tabular}{lccccccccccc}
\hline
Method & power & low\_veg & imp\_surf & car & fence\_hedge & roof & fac & shrub & tree & Avg. F1 & OA \\ \hline

DAPnet(BASE) &77.7 & 87.1 & 91.5 & 72.7 & 45.6 & 88.1 & 60.4 & 56.1 & 80.8 & 73.3 & 84.5 \\ \hline
DAPnet(M) & 81.7 & 87.2 & 93.0 & 75.3 & 50.6 & 90.7 & 63.7 & 57.9 & 82.3 & 75.8 & 86.0  \\
DAPnet(G) & 76.8 & 87.4 & 92.2 & 75.8 & 51.2 & 90.6 & 64.2 & 60.0 & 83.7 & 75.8 & 86.1  \\
DAPnet(GM) & 85.9 & 87.8 & 92.5 & 79.3 & 53.8 & 93.7 & 68.2 & 63.9 & 84.6 & 78.8 & 87.6  \\
DAPnet(P) & 85.5 & 86.9 & 93.1 & 76.8 & 51.1 & 93.5 & 69.2 & 57.4 & 86.2 & 77.8 & 87.3  \\
DAPnet(PM) & 87.4 & 88.9 & 92.7 & 81.8 & 54.4 & 94.8 & 71.4 & 58.5 & 87.2 & 79.7 & 88.3  \\ 
DAPnet(PG) & 87.6 & 90.4 & 94.7 & 85.9 & 53.4 & 96.2 & 70.3 & 67.9 & 88.2 & 81.6 & 90.2 \\  \hline
DAPnet(PGM) & 87.6 & 90.4 & 95.3 & 87.4 & 54.5 & 96.7 & 71.5 & 68.1 & 88.9 & 82.3 & 90.7  \\
 \hline
\end{tabular}}
\label{tab:abl}
\end{table*}

\subsection{Generalization Analysis}
In order to validate the generalization ability of the DAPnet, we have used the same experimental strategy as the GAPHNet. The DAPnet trained on the Vaihingen 3D dataset is directly used to predict semantic labels for the DFC 3D dataset without retraining. 
Considering that the classes of the two datasets are different, we also adopt the same strategy~\cite{LI202026} and use the trained model, mapping the predicted impervious surface to ground, shrubs, and trees to high vegetation, roofs and facades to buildings, and the rest to others. Here we mainly analyze the classification results of these three classes on the DFC 3D dataset.

We select 16 point cloud files (16,641,946 points) out of 110 files into the validation set.
We compared the experimental results with several other methods for the per-class F1 scores as shown in Table~\ref{tab:transfer}. In this transferred task, the DAPnet still obtained the best results. All of these methods use multiscale features, but the models that employ the attention mechanism have remarkable advantages in performance over those that do not. On high vegetation, both the GAPHNet and the DAPnet using the attention mechanism even outperform the other methods by more than 30\%. The incorporation of the attention module improves the generalization ability of the models. Compared to the GAPHNet, the improvement of DAPnet is not as significant as on the Vaihingen 3D dataset. We believe this is because the shape of objects in these classes is not complex, and the point clouds around object boundaries are more distinct and have fewer mix-ups between points from different classes. So even though the overall performance of the two models is close, we believe the DAPnet can be applied to more scenarios, and its best performance on the DFC 3D dataset attests to its good generalization ability.

\subsection{Ablation Study}
In order to verify the effectiveness of attention modules and the multi-radius group sampling method employed in the DAPnet, we have conducted an ablation study to compare different strategies by the incorporation or the removal of modules. These strategies build upon the model with the basic configuration (BASE), i.e., neither any attention modules are added, nor the multi-radius group sampling is used. Other strategies incorporate different modules and methods, including the PAM (P), the GAM (G), or multi-radius group sampling (M) and the combinations among them. The radius values of choice in the group sampling are in [0.1, 0.2, 0.4, 0.8]. The results are shown in Table~\ref{tab:abl}, with the strategies represented by the concatenation of incorporated module notations. The DAPnet(PGM) is the full version of our proposed model. 

Through Table~\ref{tab:abl}, we can infer the effect of different modules by comparing pairs of related strategies. For starters, the effect of the point and group attention module combined can be evidenced in the fact that the DAPnet(PG) has achieved higher overall accuracy than the DAPnet(BASE) and that the overall accuracy of DAPnet(PGM) is also higher than the DAPnet(M). By comparing DAPnet(P) or DAPnet(G) to the DAPnet(BASE), we learn that the overall accuracy of the model improves even when only one of the PAM and the GAM is added to the model. 
Whereas for the three operations, P, G, and M, the PAM has exerted a most significant improvement on the model, and their impact on the per-class F1 scores varies.
For example, for powerline, the F1 score of DAPnet(P) is higher than that of DAPnet(G), while for low vegetation and shrubs, the F1 score of DAPnet(P) is lower than that of DAPnet(G). 
The GAM is beneficial for classes with diverse shapes, while the PAM is beneficial for point clouds at the boundary areas between objects of different classes where points easily blend. However, for the overall performance of the model, the PAM attributes greater improvement than the GAM. 
To summarize, with the double attention modules, the DAPnet can obtain the best results.

\subsection{Visualization of results}

\begin{figure}[tp]
    \begin{center}
    \includegraphics[scale=1.0]{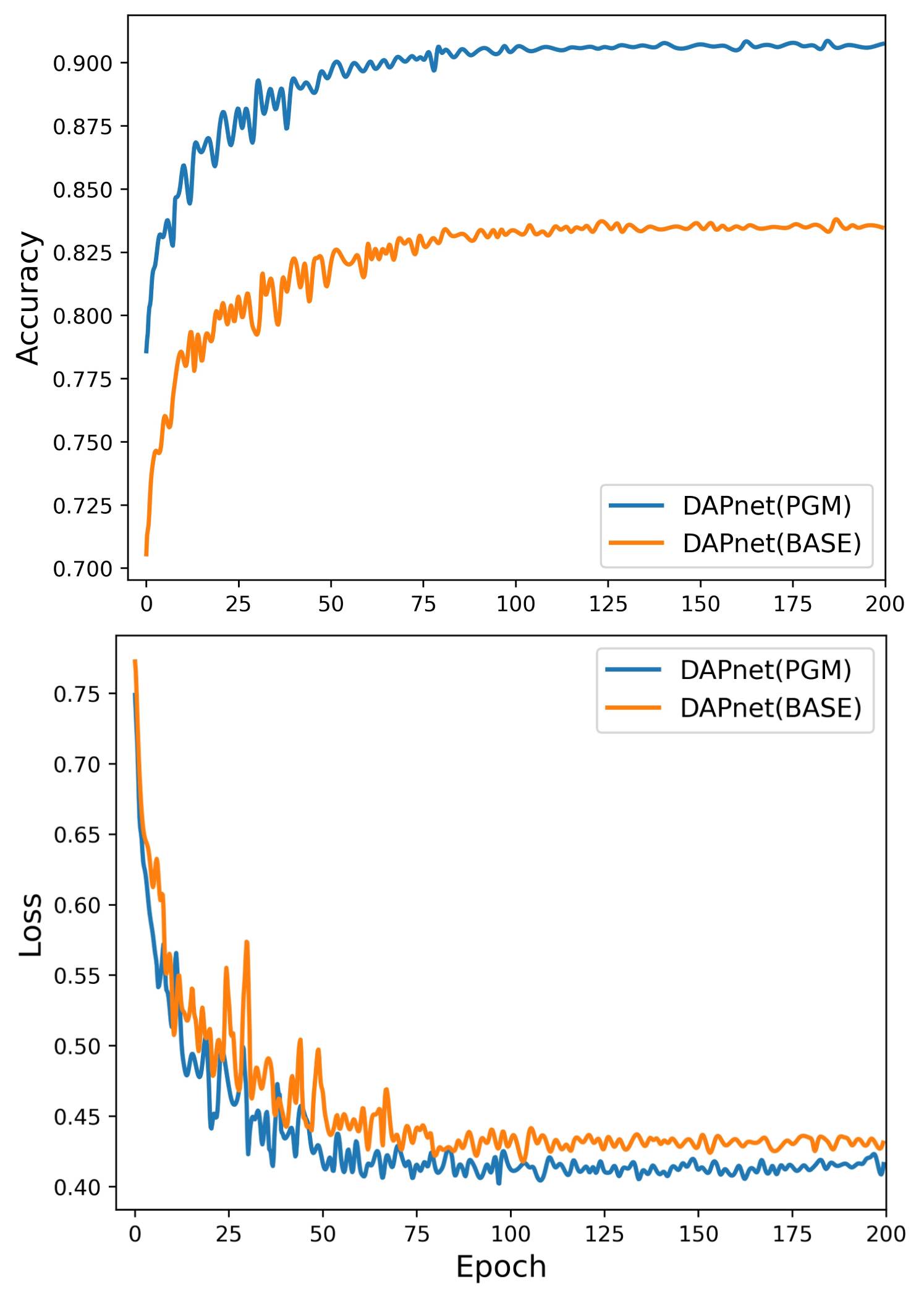}
    \end{center}
    \caption{The training phases of loss and overall accuracy from different models.}
    \label{fig:loss}
\end{figure}

In order to qualitatively analyze the performance of the models with or without the PAM and the GAM, 
Figure~\ref{fig:diff} shows the classification results of 3 random blocks in the test set under DAPnet(M) and DAPnet(PGM). These blocks contain point clouds of multiple classes, and we have marked the areas with large differences among classification results with red circles. The proposed DAPnet(PGM) both yielded good results. For the mixed areas of roofs and trees and the areas of shrubs and low vegetation, DAPnet(PGM) can classify point clouds well compared to DAPnet(M).
The correct classification of a large number of these points improves the classification performance of DAPnet, which show that DAPnet performs better with the irregular format and the border issue.
We also plot the loss and overall accuracies for DAPnet(BASE) and DAPnet(PGM), as shown in Figure~\ref{fig:loss}. We find that both DAPnet(BASE) and DAPnet(PGM) converge at approximately 100 epochs. Incorporating the double self-attention module reduces the loss and improves the overall accuracy of the model, which demonstrates the effectiveness of the double self-attention module.

\subsection{Discussion}
In the experiment, our method obtains an overall accuracy of 90.7\% using the ISPRS 3D Semantic Labeling data, which outperforms the state-of-the-art benchmark. Our model also achieves a better per-class F1 score (average +3.4\%) than the best per-class F1 scores among the benchmark methods. In particular, the accuracies of powerline and car have improved by 10.1\% and 13.2\% respectively. Through ablation experiments, we find that the two different attention modules perform differently over classes. Compared with the strategy without any attention module, the double self-attention module has significantly improved the results of various classes, especially car and shrub, which have improved by more than 10.0\%. The GAM enhances the 
F1 scores of about 3\% on the classes of car, facade, and shrub, while the PAM enhances the F1 scores by 5\% in the classes of powerline, car, facade, and tree.
We find that the marginal benefit of adding the PAM is greater than the marginal benefit of adding the GAM. When the model only uses the GAM, the model performance is similar to the model using only a multiscale radius group sampling method. However, the PAM performs best when used together with the GAM.
The ablation experiments show that the strategy of adding the double self-attention module is the best in terms of both overall accuracy and the per-class F1 scores.

\section{Conclusion}
\label{conclusion}
This study proposes a novel model called the DAPnet by using a double self-attention module to classify point cloud objects of various shapes and scales. The double self-attention module consists of point and group attention modules.
The PAM effectively solves the point cloud classification problem for multiple classes in a mixed area by assigning new weights to the extracted features through the relationship between points. The GAM, on the other hand, solves the point cloud classification problem of classes with diverse shapes through the relationship between different groups.
Under the PAM and the GAM, the DAPnet performs well and also has good generalization ability.

However, the DAPnet also has the following limitations: (1) In data preprocessing, the feature extraction of point clouds is limited by the block size and the sampling number of points. When the scale or shape of the object is larger than the block, it is difficult for the model to get good features. (2) In some areas with mixed objects from different classes, such as the roof and impervious surfaces, there are still many problems with their predicted semantic labels.
This might be because the features assigned new weights by the attention module make the model classification biased. (3) We also notice that our method and the benchmark methods have low F1 scores on the fence class. The features of these classes may be more difficult to extract than the others. 
In our future work, we would break through the limitations to get better performance.

\section*{ACKNOWLEDGEMENTS}\label{ACKNOWLEDGEMENTS}

This work was supported by the National Natural Science Foundation of China (grant numbers 41871364, 41871276, 41871302, and 41861048). This work was carried out in part using computing resources at the High Performance Computing Center of Central South University.

\bibliographystyle{IEEEtran}
\bibliography{IEEEabrv,ref}

%

\end{document}